\documentclass[nohyperref]{article}

\usepackage{microtype}
\usepackage{graphicx}
\usepackage{subfigure}
\usepackage{booktabs} 
\usepackage{color}
\usepackage{listings}

\usepackage{hyperref}
\usepackage{url}

\usepackage[frozencache,cachedir=.]{minted}
\usepackage[accepted]{icml2022}

\usepackage{amsmath}
\usepackage{amssymb}
\usepackage{mathtools}
\usepackage{amsthm}

\usepackage[capitalize,noabbrev]{cleveref}

\theoremstyle{plain}

\theoremstyle{definition}

\theoremstyle{remark}

\usepackage[textsize=tiny]{todonotes}

\icmltitlerunning{DynaMixer: A Vision MLP Architecture with Dynamic Mixing}

\begin{document}

\twocolumn[
\icmltitle{DynaMixer: A Vision MLP Architecture with Dynamic Mixing}



\icmlsetsymbol{equal}{*}

\begin{icmlauthorlist}
\icmlauthor{Ziyu Wang}{tencent}
\icmlauthor{Wenhao Jiang}{tencent}
\icmlauthor{Yiming Zhu}{tsinghua}
\icmlauthor{Li Yuan}{pku}
\icmlauthor{Yibing Song}{ailab}
\icmlauthor{Wei Liu}{tencent}
\end{icmlauthorlist}

\icmlaffiliation{tencent}{Data Platform, Tencent}
\icmlaffiliation{ailab}{Tencent AI Lab} 
\icmlaffiliation{pku}{School of Electrical and Computer Engineering, Peking University}
\icmlaffiliation{tsinghua}{Graduate school at ShenZhen, Tsinghua university}

\icmlcorrespondingauthor{Wenhao Jiang}{cswhjiang@gmail.com}
\icmlcorrespondingauthor{Wei Liu}{wl2223@columbia.edu}

\icmlkeywords{MLP, vision recognition}

\vskip 0.3in
]



\printAffiliationsAndNotice{}  

\begin{abstract}
Recently, MLP-like vision models have achieved promising performances on mainstream visual recognition tasks. In contrast with vision transformers and CNNs, the success of MLP-like models shows that simple information fusion operations among tokens and channels can yield a good representation power for deep recognition models. However, existing MLP-like models fuse tokens through static fusion operations, lacking adaptability to the contents of the tokens to be mixed. Thus, customary information fusion procedures are not effective enough. To this end, this paper presents an efficient MLP-like network architecture, dubbed DynaMixer, resorting to dynamic information fusion. Critically, we propose a procedure, on which the DynaMixer model relies, to dynamically generate mixing matrices by leveraging the contents of all the tokens to be mixed. To reduce the time complexity and improve the robustness, a dimensionality reduction technique and a multi-segment fusion mechanism are adopted. Our proposed DynaMixer model (97M parameters) achieves 84.3\% top-1 accuracy on the ImageNet-1K dataset without extra training data, performing favorably against the state-of-the-art vision MLP models. When the number of parameters is reduced to 26M, it still achieves 82.7\% top-1 accuracy,  surpassing the existing MLP-like models with a similar capacity. The code is available at \url{https://github.com/ziyuwwang/DynaMixer}.
\end{abstract}

\begin{figure}[!t]
\begin{center}
\includegraphics[width=1.0\columnwidth]{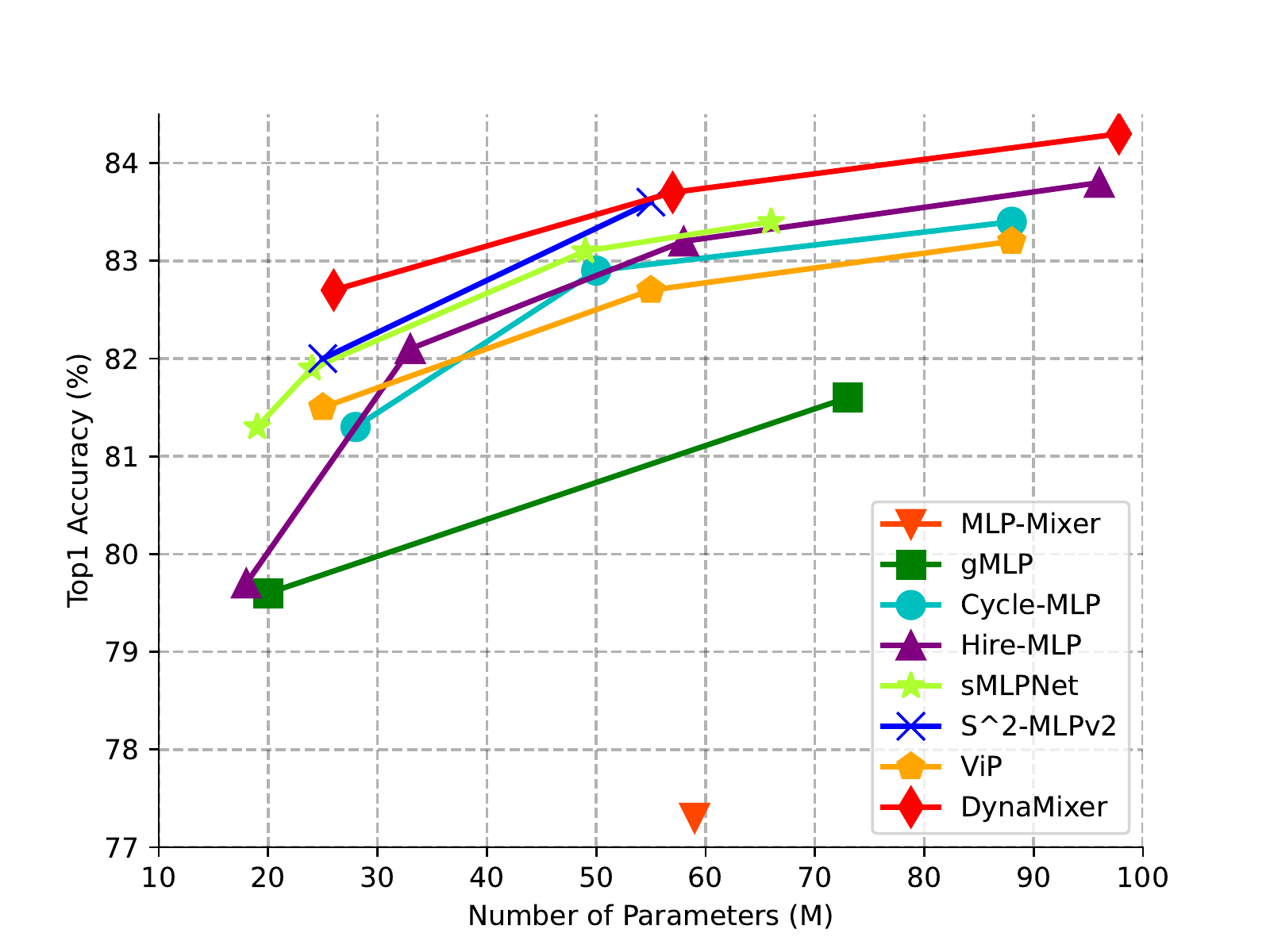}
\end{center}
\caption{
\textbf{ImageNet accuracy v.s. model capacity.} 
All models are trained on ImageNet-1K without extra data. DynaMixer outperforms all the existing MLP-like models, such as MLP-Mixer~\cite{mlp-mixer}, Hire-MLP~\cite{hire-mlp}, $S^2$-MLP-v2~\cite{s2-mlp}, ViP~\cite{vip}, and Cycle-MLP~\cite{cyclemlp}.
}
\label{fig:acc_cap}
\end{figure}

\section{Introduction}
Convolutional neural networks (CNNs) have been the dominating solution for a wide range of computer vision tasks for a long time, \textit{e.g.,} visual recognition and segmentation. 
{CNNs aim at training the whole network end-to-end to remove hand-crafted visual features and inductive biases.}
And they heavily rely on convolution operators and pooling operators, which introduce locality and spatial invariances. Recently, Transformers~\cite{vaswani2017attention}, which have achieved overwhelming success in natural language processing, have also been introduced into the compute vision field. The Vision Transformers (ViTs)~\cite{vit} have achieved state-of-the-art performances on visual recognition tasks. In ViT, images are first divided into a sequence of non-overlapping patches, which can be seen as word tokens in the natural language processing field. Those patches are then fed into a stack of blocks based on self-attention to obtain the contextualized output tokens. ViT removes the convolution and pooling operators from CNNs, and employs a self-attention mechanism to model the relationships among image patches. 
Inspired by ViTs, many efforts have been made to designing simpler models in the setting with a large amount of data available.


More recently, a pure multilayer perceptron (MLP) architecture~\cite{mlp-mixer}, called MLP-Mixer, was proposed to reduce the inductive biases further and showed competitive performance with CNN-based models and vision transformers. MLP-Mixer is much simpler than transformer-based models because it utilizes MLP as a building block, removing the need of invoking self-attention. In MLP-Mixer, each layer mainly relies on two steps to perform information interaction: token mixing step and channel mixing step, both based on MLP blocks. The token mixing step performs fusion among the tokens, while the channel mixing step performs fusion on the channel dimension.
Inspired by this, a number of MLP-like models~\cite{cyclemlp,vip,s2-mlp,s2-mlpv2} have emerged for further improvements. 
However, these MLP-like methods rely on fixed static mixing matrices for patch communications, which may restrict the adaptability to the contents to be fused.

To solve the above limitation, we propose a vision MLP architecture with dynamic mixing, dubbed DynaMixer, which can generate mixing matrices dynamically for each set of tokens to be mixed by considering their contents. Note that mixing all the image tokens consumes a significant time cost. For a computational speedup, we mix tokens in a row-wise and column-wise way. During each mixing process, we reduce the feature dimensionality to generate the mixing matrices. We empirically find that we can reduce feature dimensionality significantly without undermining the performance much. \textit{The feature dimensionality can even be reduced to 1 in our experiments.}  Furthermore, we divide the feature channels into multiple segments and perform token mixing separately. This improves the mixing robustness as well as efficiency. As a result, our method achieves state-of-the-art performance among existing MLP-like vision models on the ImageNet-1K dataset~\cite{imagenet}, as summarized in Fig~\ref{fig:acc_cap}.  Without any extra data, our DynaMixer accomplishes 84.3\% top-1 accuracy with 97M parameters. When the model parameters are reduced to 26M, DynaMixer can still reach 82.7\% top-1 accuracy, far exceeding other MLP-like models with a similar number of parameters.

\section{Related Work}
The deep neural networks for vision related tasks evolve from CNNs to ViTs and MLPs. In this section, we review these models and illustrate the relationships between our DynaMixer and these models.

\subsection{Convolutional Neural Networks}
The CNNs have drawn huge attention since AlexNet~\cite{alexnet} improved image classification by a large margin. Convolution operations, nonlinear activation, and pooling operations have become prevalent for CNN architecture designs. 
Based on the previous works on CNN, inceptions~\cite{szegedy2015going,szegedy2016rethinking,szegedy2017inception} were proposed by employing multiple parallel branches. 
A breakthrough was made by ResNet~\cite{resnet}, in which the network depth was significantly increased. To ease training, skip connections with identity mapping were introduced into ResNets. They achieved state-of-the-art visual recognition performances at that time and inspired further research, including ResNeXt~\cite{xie2017aggregated} and DenseNet~\cite{huang2017densely}. Afterward, attention mechanisms were gradually explored to benefit CNNs. Examples include SENet~\cite{hu2018squeeze}, non-local neural network~\cite{wang2018non}, and local relationship network~\cite{hu2019local}.
CNNs have been widely applied to computer vision related tasks, such as visual recognition, image/video generation, object detection, semantic segmentation, and so on.



\subsection{Vision Transformers}

Transformer~\cite{vaswani2017attention} was firstly proposed in the machine translation area. 
It was then adopted for computer vision applications in ViT~\cite{vit}. Input images in ViT are divided into a sequence of non-overlapping patches, which are regarded as tokens and are fed into transformers. 
Unlike CNNs aggregating information within local windows, ViTs model global information interactions among visual tokens via self-attention. Inspired by ViTs, many variants are proposed to remove the drawbacks of original ViTs.
In DeepViT~\cite{zhou2021deepvit} and CaiT~\cite{cait}, the problem that the performance of ViTs saturates fast when scaled to be deeper was addressed. In~\cite{yuan2021volo,chen2021crossvit,han2021transformer,swin,crossformer,fan2021multiscale,ge2021revitalizing,liang2022not,tong2022videomae,bai2022improving,zhou2021refiner}, multi-granularity was introduced into ViTs to improve the generalization ability. A complete survey on vision transformers can be found in~\cite{han2020survey,khan2021transformers}. 
All transformer-based models benefit from the self-attention mechanism for its flexibility and adaptiveness. 
Our model does not involve the self-attention mechanism while maintaining adaptability.
Differences will be discussed in the following section.



 \begin{figure*}[ht]
\begin{center}
\includegraphics[width=.94\linewidth]{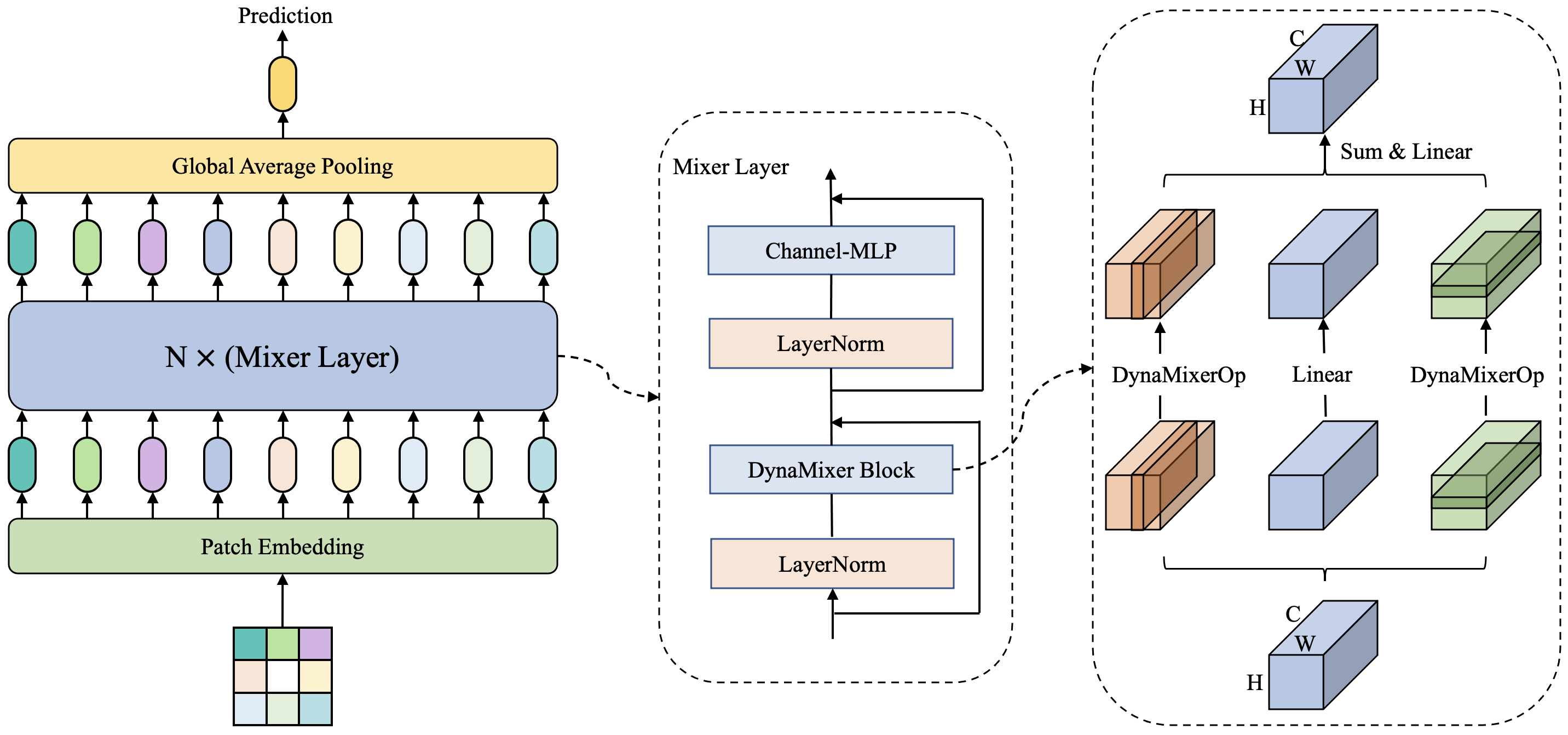}
\end{center}
\caption{
\textbf{The basic architecture of our proposed DynaMixer.}
 DynaMixer consists of a patch embedding layer, several mixer layers, a global average pooling layer, and a classifier head. The patch embedding layer transforms  input non-overlapping patches into corresponding input tokens, which are fed into a sequence of mixer layers to generate the output tokens. All output tokens are averaged in the average pooling layer, and the final prediction is generated with a classifier head. The mixer layer (middle part) contains two layer-normalization layers, a DynaMixer block and a channel-MLP block. The DynaMixer block (right part) performs row mixing and column mixing via the DynaMixer operations (depicted in Fig.~\ref{fig:acc_dyna_mixer_op}), and a simple channel-mixing via linear projection. The mixing results are summed and outputted with a linear transformation. 
}
\label{fig:overview}
\end{figure*}       

\subsection{Vision MLPs}
Recently, MLP-based vision models~\cite{mlp-mixer,resmlp} were proposed to reduce the inductive bias further and showed competitive performances with CNN-based models and ViTs. 
Existing MLP-like models share a similar macro framework but have different micro block designs. MLP-like models usually divide one input image into patches, like in vision transformers, and then perform two main steps: token mixing step and channel mixing step. The specific details of these two steps, especially token-mixing steps, are different among the existing methods.
ViP~\cite{vip} was proposed to perform token mixing along the height and width dimensions with simple linear projections to encode spatial information, which is different from other MLP-like models that fuse token information among all tokens in one step. 
S$^2$-MLP~\cite{s2-mlp} replaced the token mixing step with a spatial-shift step to enable information interaction among tokens. 
CycleMLP~\cite{cyclemlp} samples local points along the channel dimension in a cyclical style, while AS-MLP~\cite{asmlp} samples locations in a cross. Hire-MLP~\cite{hire-mlp} mixes tokens within a local region and across local regions.


 All these MLP-like methods
 rely on static mixing matrices for patch communications, which may limit the performance for lacking adaptiveness since it is natural that a different set of tokens should be mixed differently. 
 Thus, we propose to generate the mixing matrices dynamically by considering the contents of tokens to be mixed.
 A recent work, Synthesizer~\cite{synthesizer}, pointed out that either the attention matrix in ViTs or the mixing matrix in MLP can be seen as generated with a specially designed function. It is worth noting that although the Synthesizer~(Dense)~\cite{synthesizer} adopts to generate the mixing matrix dynamically, the mixing weights for a specific position are determined only by the contents of the token at the same position. However, the weights for it in our model are determined by all the tokens. We will show how to design such a generating function and how to reduce the computational complexity in the following section.


\section{The Architecture of DynaMixer}

In this section, we present our model designs. The overall framework of our network is illustrated in Fig.~\ref{fig:overview}. Like in other MLP-based models,  the input image is divided into non-overlapping patches. All patches are projected onto the embedding space with a shared matrix, which are then fed into a sequence of mixer layers to generate the output tokens. All output tokens are averaged in the average pooling layer, and then sent into the classifier head to yield the final prediction.
Except for the layer-normalization layers and skip connections, the mixer layer contains a DynaMixer block and a channel-MLP block, which are responsible for fusing token information and channel information, respectively. The channel-MLP block is just a feed-forward layer as in Transformer~\cite{vaswani2017attention}.
The DynaMixer block performs row mixing and column mixing via DynaMixer operations, and channel-mixing via simple linear projection, respectively. The mixing results are aggregated and outputted via a linear transformation. 
We first illustrate our DynaMixer operation, and then elaborate on the DynaMixer block. Finally, we analyze the differences from other MLP-like models. 

\subsection{The DynaMixer Operation}

\begin{figure}[t]
\begin{center}
\includegraphics[width=0.6\columnwidth]{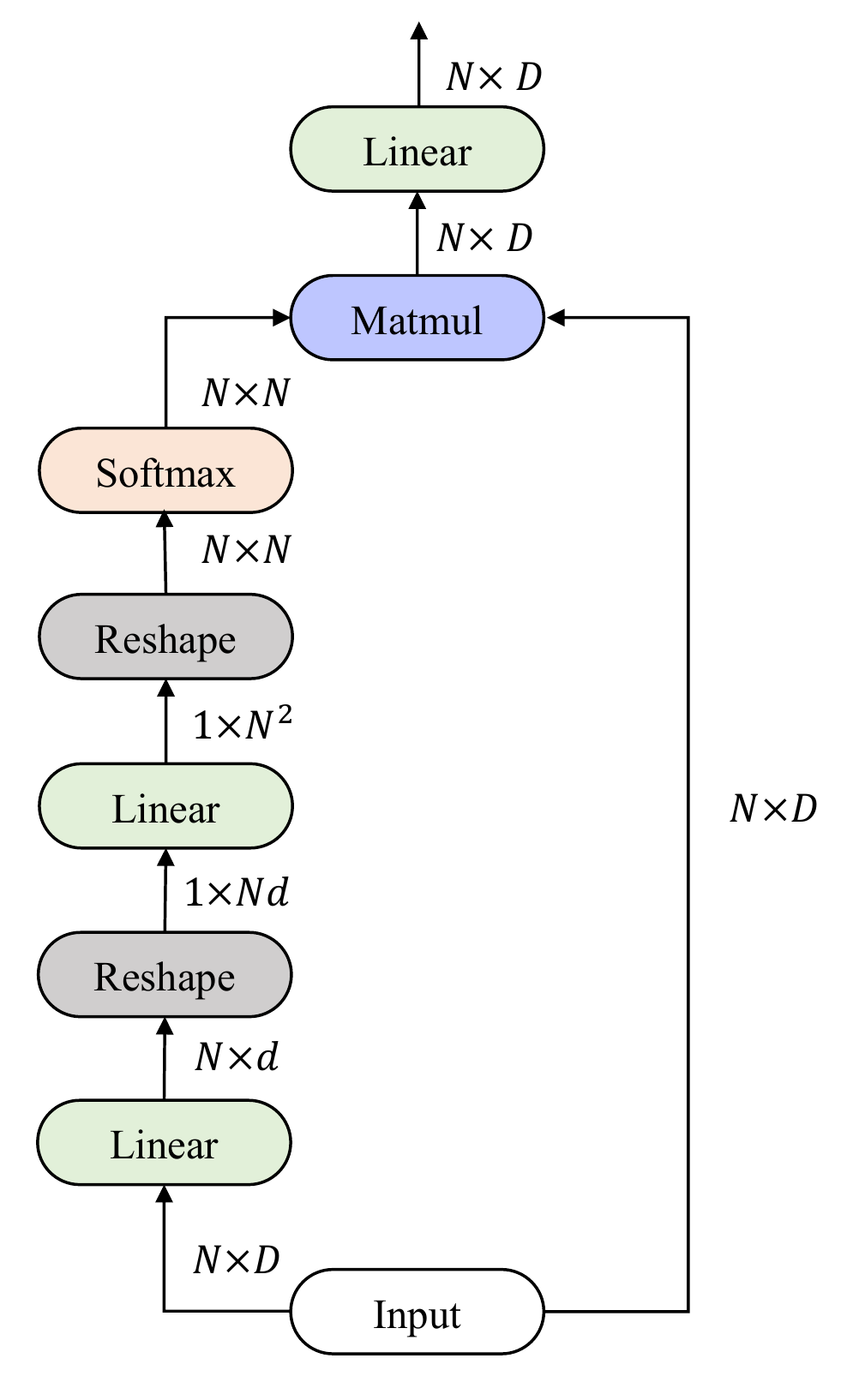}
\end{center}
\caption{The procedure of our proposed DynaMixer operation for one segment.
}
\label{fig:acc_dyna_mixer_op}
\end{figure}

\paragraph{Dynamic mixing matrix generation.}
The principle of our design is to generate a dynamic mixing matrix $P$ given a set of input tokens $X \in R^{N \times D}$ by considering their contents, where $N$ is the number of tokens, and $D$ is the feature dimensionality. Once we obtain $P$, we can mix the tokens by $Y = PX$ to obtain the output tokens $Y$. 
A simple way to get $P$ is utilizing a linear function of all input token features to estimate it. Thus, we can simply flatten $X$ into a vector and generate the mixing matrix as
\begin{align}
P_{i\cdot} = \texttt{softmax} \left( \texttt{flat}(X)^T W^{(i)} \right),
\end{align}
where $\texttt{flat}(X) \in R^{ND \times 1}$ is a vector by flattening $X$, 
$\texttt{softmax}(\cdot)$ is the softmax operator performing on a row vector,
$W^{(i)} \in R^{ND\times N}$, and $P$ is the mixing matrix. $P_{i\cdot}$ is the $i$-th row of $P$, and it contains the mixing weights for the $i$-th output token. However, the number of parameters of the above process is too large since $ND$ is usually too big. Thus, we can perform dimensionality reduction first to reduce the number of parameters. Therefore, we have the following steps to generate the mixing matrix and perform mixing on tokens:
\begin{align}
\hat{X} &= X W_d, \\
P_{i\cdot} &= \texttt{softmax} \left(\texttt{flat}(\hat{X}) W^{(i)} \right),  \\
Y &= P X,
\end{align}
where $ \hat{X} \in R^{N \times d}$, and $d \ll D$ is a quite small number, say 1 or 2. The whole procedure for generation and mixing is described pictorially in Fig~\ref{fig:acc_dyna_mixer_op}, which can be implemented easily with PyTorch~\cite{pytorch} or TensorFlow~\cite{tensorflow2015-whitepaper}.

\paragraph{Multi-segment fusion mechanism.}  To improve the robustness of our model, we divide features into $S$ segments, perform the mixing operation separately, and combine the mixed results to obtain the final results:
\begin{align}
\hat{X}^{(s)} &= X W_d^{(s)}, \\
P^{(s)}_{i\cdot} &= \texttt{softmax} \left(\texttt{flat}(\hat{X}^{(s)}) W^{(s,i)} \right), \label{eq_dyna_mix_op_w} \\
Y &= [P^{(0)} X^{(0)}, \cdots, P^{(S-1)} X^{(S-1)}] W_o,
\end{align}
where $\hat{X}^{(s)} \in R^{N \times d}, W^{(s, i)} \in R^{Nd \times N}$, $P^{(s)}_{i\cdot}$ is the $i$-th row of the token mixing matrix for the $s$-th segment,
$X^{(s)}$ is the tokens in the $s$-th segment,
$[\cdot, \cdot]$ is the concatenation operation, and $W_o \in R^{D \times D}$ is a feature fusion matrix for output.
Note that,  different dimensionality reduction matrices, \textit{i.e., $W_d^{(s)}$}, are used for different segments, which are beneficial for the expressive power of the model. Moreover, the mixing matrix for the $s$-th segment is also effected by the contents of other segments. Thus, the mixing behaviours for different segments are not independent.

\paragraph{Weights sharing among segments.}
To reduce the number of parameters,  the matrices $W^{(s,i)}$ are shared among all segments. Thus, the number of parameters of the DynaMixer operation is $S \times D\times d +  N^3 \times d + D^2$. In practice, we found that $d$ can be quite small, \textit{e.g.}, $d=1$ or $d=2$.

We use the following expression to denote the DynaMixer operation in the rest of this paper:
\begin{align}
    Y = \texttt{DynaMixerOp}(X).
\end{align}




    
    
    

\subsection{The DynaMixer Block}

The number of parameters of the above operation mainly depends on $N$, the number of tokens to be mixed, which is $H \times W$, where $H$ and $W$ are the numbers of patches in the height and width directions, respectively. Thus, to reduce the computational complexity, we adopt a strategy similar to ViP~\cite{vip}, which mixes tokens in a row or column at one step. As depicted in the right part of Fig~\ref{fig:overview}, the DynaMixer block consists of three components, which are row mixing, column mixing, and channel mixing. In row mixing, the tokens that belong to the same row are mixed via DynaMixer operations. And the parameters of DynaMixer operations are shared among all rows. Similarly, the tokens that belong to the same column are mixed via DynaMixer operation in column mixing, in which the parameters of DynaMixer operations are also shared. The channel mixing is simply a linear transformation on features. We denote the outputs of the three components as $Y_h$, $Y_w$ and $Y_c$, and the output of the DynaMixer block is
\begin{align}\label{dyna_mixer_sum}
    Y_{out} = (Y_h + Y_w + Y_c)W'_o.
\end{align}
The procedure for the DynaMixer block is summarized in Algorithm~\ref{alg_dyna_mixer_block}, in which we denote the input as $X \in R^{H \times W \times D} $ for convenience.





\begin{algorithm}[t]
	\caption{Pseudo-code for DynaMixer Block (PyTorch-like)}
\label{alg_dyna_mixer_block}
\begin{minted}[fontsize=\small]{python}
###### initializaiton #######
proj_c = nn.Linear(D, D)
proj_o = nn.Linear(D, D)
    
###### code in forward ######
def dyna_mixer_block(self, X):
  H, W, D = X.shape
    
  # row mixing
  for h = 1:H
    Y_h[h,:,:] = DynaMixerOp_h(X[h,:,:])
        
  # column mixing
  for w = 1:W
    Y_w[:,w,:] = DynaMixerOp_w(X[:,w,:])
    
  # channel mixing
  Y_c = proj_c(X)
  Y_out = Y_h + Y_w + Y_c
    
  return proj_o(Y_out)
\end{minted}
\end{algorithm}

    
    
        
    
    





In Algorithm~\ref{alg_dyna_mixer_block}, the three mixing results are just summed to obtain the final result, as described in Eq.~\eqref{dyna_mixer_sum}. To further improve the performance, importances of the three components can be estimated by a method similar to~\cite{split-atten,vip}. In the following experiments, the weighted DynaMixer block is used by default.

\subsection{Discussion }
\paragraph{DynaMixer v.s. Synthesizer (Dense).}
The weights for the $i$-th output token generated by Synthesizer~(Dense) are only determined by the contents of the $i$-th input token. Thus, \textit{it cannot model the effects from the other tokens, leading to inaccurate weight estimation.} However, our model considers all the tokens to generate more accurate mixing weights.
Moreover, our model divides the channel into non-overlapping segments and performs separate mixing operations on them. Such a design can improve the robustness and expressive power of our model. 

\paragraph{DynaMixer v.s. ViP.}
Except that the mixing matrices in our model are generated dynamically, the mixing behaviors are also quite different. Suppose $W$ tokens in the $h$-th row are to be mixed. Each token is divided into $S$ segments along channel dimension, and the dimensionality of channel is denoted as $D$. In ViP, \textit{$W\times S$ vectors of dimensionality $\frac{D}{S}$ will be mixed by multiplying one fixed matrix}. However, in DynaMixer, the mixing operation occurs $S$ times. One time for one segment. And \textit{each mixing operation is performed on $W$ vectors of dimensionality $\frac{D}{S}$ with different mixing matrices}. 

\paragraph{DynaMixer v.s. self-attention.}
The mixing matrix (attention map) in self-attention models the \textit{pair-wise relationships between tokens}. Thus, it is a quadratic function of token features. However, in our method, the mixing matrix is estimated by computing the \textit{relationships between tokens and a trainable matrix} ($W^{s,i}$ in Eq.~\eqref{eq_dyna_mix_op_w}).
\section{Experimental Results}
In this section, we present the experimental results and analysis. First, we will give the configurations of our DynaMixer used in the experiments, and then the experimental settings and results on the ImageNet-1K dataset are provided. At last, the ablation studies are presented to provide a deep understanding of the designs in our model.

\subsection{Configurations of DynaMixer}
The configurations of our model used in the experiments are summarized in Table~\ref{tab:dyna_mixer_config}. There are three versions of DynaMixer, denoted as ``DynaMixer-S'', ``DynaMixer-M'', and ``DynaMixer-L'', according to the model sizes. In all our experiments, the input image size is $224 \times 224$, and the input patch size is $7 \times 7$. All our models have two stages, and each starts with a patch embedding layer. The patch size for the second stage is $2 \times 2$. Thus, the corresponding patch embedding layer can be seen as a downsampling layer.

\begin{table}[ht]
    \resizebox{.99\columnwidth}{!}{
    \centering
    \begin{tabular}[t]{lccc}
    \toprule
    Specification & DynaMixer-S & DynaMixer-M & DynaMixer-L\\
    \midrule
    Patch size & $7 \times 7$ & $7 \times 7$ & $7 \times 7$ \\
    Hidden size & 192 & 256 & 256 \\
    \#Tokens & $32 \times 32$ & $32 \times 32$ & $32 \times 32$ \\
    \#Mixer Layers & 4 & 7 & 8 \\
    $S$ & 8 &8  &8  \\
    \midrule
    Patch size & $2 \times 2$ & $2 \times 2$ & $2 \times 2$ \\
    Hidden size & 384 & 512 & 512 \\
    \#Tokens & $16 \times 16$ & $16 \times 16$ & $16 \times 16$ \\
    \#Mixer Layers & 14 & 17 & 28 \\
    $S$ &16  & 16 & 16 \\
    \midrule
    \#Layers & 18 & 24 & 36 \\
    MLP Ratio & 3 & 3 & 3 \\
    Stoch. Dep. & $0.1$ & $0.1$ & $0.3$ \\
    $d$ & 2 & 2 & 8 \\
    \#Parameters & 26M & 57M & 97M \\
    \bottomrule
    \end{tabular}
    }
    \caption{\textbf{The configurations of DynaMixers used.} We design three  DynaMixers (DynaMixer-S, DynaMixer-M, and DynaMixer-L) of different sizes (Small, Medium, and Large) for the experiments, according to the number of parameters. 
    The first two groups of rows show the configurations of the first and the second stage of the models, respectively.
    The third group of rows shows some other hyperparameters.
    }
    \label{tab:dyna_mixer_config}
\end{table}

\subsection{Experimental Configurations}
We train our proposed DynaMixer on the public image classification benchmark ImageNet-1K dataset~\cite{imagenet}, which covers $1K$ categories of natural images and contains $1.2M$ training images and $50K$ validation images. Because the test set for this benchmark is unlabeled, we follow the common practice by evaluating the performance on the validation set. The code implementation is based on PyTorch~\cite{pytorch} and 
the TIMM\footnote{https://github.com/rwightman/pytorch-image-models} toolbox.

We train our model with one machine with 8 NVIDIA A100 GPUs with data parallelism.
For model optimization, we adopt the AdamW optimizer~\cite{loshchilov2017decoupled}.
The learning rates for DynaMixer-S, DynaMixer-M, and DynaMixer-L are 0.002,
and the corresponding batch sizes on one GPU are 256, 128, and 64, respectively.
We set the weight decay rate to 0.05 and set the warmup learning rate to $10^{-6}$ to follow the settings in previous work~\cite{touvron2021training,token-labeling}. We use automatic mixed-precision of the PyTorch version for training acceleration. Stochastic Depth~\cite{huang2016deep} is used, and the drop rate varies with the size of the model. The model is trained for 300 epochs. For data augmentation methods, we use CutOut~\cite{zhong2020random}, RandAug~\cite{randaugment}, MixUp~\cite{mixup}, and CutMix~\cite{yun2019cutmix}. Detailed settings about $S$, $d$, and the path drop rates for different versions of DynaMixer can be found in Table~\ref{tab:dyna_mixer_config}.

\subsection{Results on ImageNet-1K}
In this subsection, we compare our model with MLP-like models, CNN-based models, and transformer-based models to show the advantages of our dynamic mixing mechanism. 

\begin{table}[ht]
\resizebox{.99\columnwidth}{!}{
    \centering
    \begin{tabular}{l|c|c|c|c}
    \toprule
    Model & Date & Param & FLOPs & Top-1 (\%) \\
    \midrule
    Mixer-B/16~\cite{mlp-mixer} &  MAY, 2021  & 59M & 12.7G & 76.4 \\
    Mixer-B/16$^{\dagger}$~\cite{mlp-mixer}  & & 59M & 12.7G & 77.3 \\
    \midrule
    ResMLP-S12~\cite{resmlp} & & 15M & 3.0G & 76.6 \\
    ResMLP-S24~\cite{resmlp} & MAY, 2021 & 30M & 6.0G & 79.4 \\
    ResMLP-B24~\cite{resmlp}  & & 116M & 23.0G & 81.0 \\
    \midrule
    gMLP-Ti~\cite{gmlp} & &6M & 1.4G & 72.3 \\
    gMLP-S~\cite{gmlp} & MAY, 2021 & 20M & 4.5G & 79.6 \\
    gMLP-B~\cite{gmlp} & & 73M & 15.8G & 81.6 \\
    \midrule
    ViP-Small/14~\cite{vip} & & 30M & 6.9G & 80.7 \\
    ViP-Small/7~\cite{vip} &  JUN, 2021 & 25M & 6.9G & 81.5 \\
    ViP-Medium/7~\cite{vip} & & 55M & 16.3G & 82.7 \\
    ViP-Large/7~\cite{vip} & & 88M & 24.4G & 83.2 \\
    \midrule
    AS-MLP-T~\cite{asmlp} & & 28M & 4.4G & 81.3 \\
    AS-MLP-S~\cite{asmlp} & JUL, 2021 & 50M & 8.5G & 83.1 \\
    AS-MLP-B~\cite{asmlp} & & 88M & 15.2G & 83.3 \\
    \midrule
    S$^2$-MLPv2-Small/7~\cite{s2-mlpv2} & AUG, 2021 & 25M & 6.9G & 82.0 \\
    S$^2$-MLPv2-Medium/7~\cite{s2-mlpv2} & & 55M & 16.3G & 83.6 \\
    \midrule
    RaftMLP-36~\cite{raftmlp} & AUG, 2021 & 44M & 9.0G & 76.9 \\
    RaftMLP-12~\cite{raftmlp} & & 58M & 12.0G & 78.0 \\
    \midrule
    sMLPNet-T~\cite{smlpnet} & SEP, 2021 & 24M & 5.0G & 81.9 \\
    sMLPNet-S~\cite{smlpnet} & & 49M & 10.3G & 83.1 \\
    sMLPNet-B~\cite{smlpnet} & & 66M & 14.0G & 83.3 \\
    \midrule
    ConvMLP-S~\cite{convmlp} & & 9M & 2.4G & 76.8 \\
    ConvMLP-M~\cite{convmlp} & SEP, 2021 & 17M & 3.9G & 79.0 \\
    ConvMLP-L~\cite{convmlp} & & 43M & 9.9G & 80.2 \\
    \midrule
    CycleMLP-T~\cite{cyclemlp} & &28M &4.4G &81.3 \\
    CycleMLP-S~\cite{cyclemlp} & NOV, 2021 &50M &8.5G &82.9 \\
    CycleMLP-B~\cite{cyclemlp} & &88M &15.2G &83.4 \\
    \midrule
    Hire-MLP-Ti~\cite{hire-mlp} & &18M & 2.1G &79.7\\
    Hire-MLP-S~\cite{hire-mlp} & NOV, 2021 &33M & 4.2G &82.1\\
    Hire-MLP-B~\cite{hire-mlp} & &58M & 8.1G &83.2\\
    Hire-MLP-L~\cite{hire-mlp} & &96M & 13.4G &83.8\\
    \midrule
    DynaMixer-S & & 26M & 7.3G & 82.7 \\
    DynaMixer-M & & 57M & 17.0G & 83.7 \\
    DynaMixer-L & & 97M & 27.4G & 84.3\\
    \bottomrule
    \end{tabular}}
    \caption{Image classification results of our DynamMixer and other MLP-like models on the ImageNet-1K benchmark without extra data. ``Top-1'' denotes Top-1 accuracy. ``Param'' and FLOPs denote the number of parameters and the number of floating point operations, respectively. Model with $^{\dagger} $ was re-implemented by ~\cite{gmlp}. The date column means the corresponding initial release date on arXiv. 
    The FLOPs are computed with the code of CycleMLP~\cite{cyclemlp}.
    }
    \label{tab:acc_mlps}
\end{table}

\paragraph{Comparisons with MLP-like Models.} First, we compare our DynaMixer with MLP-like models. The top-1 accuracies of different MLP-like models on ImageNet-1K~\cite{imagenet} without external data are shown in Table~\ref{tab:acc_mlps}. 
The ``Param'' and ``FLOPs'' columns denote the number of parameters and the number of floating point operations, respectively. The results are extracted from the corresponding papers, and the FLOPs are computed with the code of CycleMLP\footnote{https://github.com/ShoufaChen/CycleMLP}.  Firstly, it is obvious that our model consistently outperforms the other models of a similar number of parameters.
In addition, we can see that our DynaMixer-S model with 26M parameters achieves top-1 accuracy of 82.7$\%$, which has already surpassed most of the existing MLP-like models of all sizes and is even 
better than gMLP-B~\cite{gmlp} with 73M parameters. Increasing the number of parameters to 57M,  our DynamMixer-M obtains accuracy 83.7$\%$,  which is superior to all MLP-like models. Further expanding the network to 97M parameters,  DynaMixer-L can achieve top-1 accuracy of 84.3$\%$, which is a new state-of-the-art among the MLP-like models. 

By comparing with  ViP~\cite{vip}, which is the most similar model to ours, we find that  the number of parameters and FLOPs of our model are only slightly larger than those of ViP.
However, the improvements of performance are not caused by the increment on model size. The model sizes of our model and ViP are similar. For example, the number of parameters and FLOPs of DynaMixer-M are 57M and 17G. And those of ViP-Mediam/7 are 55M and 16.3G. Thus, the model size of ours is about 5\% larger than ViP. But the performance is improved greatly from 82.7\% to 83.7\%. Therefore, the model structure should be attributed to such improvements.
Specifically, the designs of dynamic mixing matrices and multi-segment fusion mechanism do help improve the performance. The mixing matrices generated dynamically could capture the contents of tokens to be mixed. Thus, different sets of tokens will be mixed with different mixing matrices. Therefore, token information could be fused more effectively. At last, the multi-segment fusion mechanism improves the robustness of the whole model. We also provide detailed ablation studies of these two designs in the following subsection for gaining a deeper understanding of our method.


\paragraph{Comparison with SOTA Classification Models.} We also compare our model with well-known state-of-the-art models, which include CNN-based models, transformer-based models, and hybrid models. The results are shown in Table~\ref{tab:acc_all}. We can see that our proposed models still achieve the best performance among models with a similar number of parameters. Specifically, our models achieve better performance than Swin Transformer~\cite{swin}, which is the state-of-the-art transformer-based model. Our DynaMixer-S achieves accuracy 82.7\% with slightly fewer parameters, while Swin-T achieved 81.3\%. For medium-sized and large-sized models, our models are still better than Swin-S and Swin-B. Our model is also better than CrossFormer~\cite{crossformer}, which employs a multi-scale strategy. We believe that by introducing multi-granularity, the performance of our method could be further improved.
From the above comparisons, we can see that DynaMixer is a promising architecture for visual recognition tasks. 


\begin{table}[ht]
\resizebox{.99\columnwidth}{!}{
    \centering
    \begin{tabular}{l|c|c|c|c|c}
    \toprule
    Model & Family & Scale & Param & FLOPs & Top-1 $(\%)$ \\
    \midrule
    ResNet18~\cite{resnet} & CNN & $224^{2}$ & 12M & 1.8G & 69.8 \\
    EffNet-B3~\cite{efficientnet} & CNN & $300^{2}$ & 12M & 1.8G & 81.6 \\
    PVT-T~\cite{pvt} & Trans & $224^{2}$ & 13M & 1.9G & 75.1 \\
    GFNet-H-Ti~\cite{gfnet} & FFT & $224^{2}$ & 15M & 2.0G & 80.1 \\
    \midrule
    ResNet50~\cite{resnet} & CNN & $224^{2}$ & 26M & 4.1G & 78.5 \\
    RegNetY-4G~\cite{regnet} & CNN & $224^{2}$ & 21M & 4.0G & 80.0 \\
    DeiT-S~\cite{touvron2021training} & Trans & $224^{2}$ & 22M & 4.6G & 79.8 \\
    BoT-S1-50~\cite{bot} & Hybrid & $224^{2}$ & 21M & 4.3G & 79.1 \\
    PVT-S~\cite{pvt} & Trans & $224^{2}$ & 25M & 3.8G & 79.8 \\
    T2T-14~\cite{Yuan_2021_ICCV} & Trans & $224^{2}$ & 22M & 4.8G & 81.5 \\
    Swin-T~\cite{swin} & Trans & $224^{2}$ & 29M & 4.5G & 81.3 \\
    GFNet-H-S~\cite{gfnet} & FFT & $224^{2}$ & 32M & 4.5G & 81.5 \\
    CrossFormer-T~\cite{crossformer} & Trans & $224^{2}$ & 27M & 2.9G & 81.5 \\
    CrossFormer-S~\cite{crossformer} & Trans & $224^{2}$ & 30M & 4.9G & 82.5 \\
    DynaMixer-S & MLP & $224^{2}$ & 26M & 7.3G & \textbf{82.7} \\
    \midrule
    ResNet101~\cite{resnet} & CNN & $224^{2}$ & 45M & 7.9G & 79.8 \\
    RegNetY-8G~\cite{regnet} & CNN & $224^{2}$ & 39M & 8.0G & 81.7 \\
    BoT-S1-59~\cite{bot} & Hybrid & $224^{2}$ & 34M & 7.3G & 81.7 \\
    PVT-M~\cite{pvt} & Trans & $224^{2}$ & 44M & 6.7G & 81.2\\
    GFNet-H-B~\cite{gfnet} & FFT & $224^{2}$ & 54M & 8.4G & 82.9 \\
    Swin-S~\cite{swin} & Trans & $224^{2}$ & 50M & 8.7G & 83.0 \\
    PVT-L~\cite{pvt} & Trans & $224^{2}$ & 61M & 9.8G & 81.7 \\
    CrossFormer-B~\cite{crossformer} & Trans & $224^{2}$ & 52M & 9.2G & 83.4 \\
    T2T-24~\cite{Yuan_2021_ICCV} & Trans & $224^{2}$ & 64M &  13.8G & 82.1 \\
    DynaMixer-M & MLP & $224^{2}$ & 57M & 17.0G & \textbf{83.7} \\
    \midrule
     CrossFormer-L~\cite{crossformer} & Trans & $224^{2}$ & 92M & 16.1G & 84.0 \\
    Swin-B~\cite{swin} & Trans & $224^{2}$ & 88M & 15.4G & 83.3 \\
    DeiT-B~\cite{touvron2021training} & Trans & $224^{2}$ & 86M & 17.5G & 81.8 \\
    DeiT-B~\cite{touvron2021training} & Trans & $384^{2}$ & 86M & 55.4G & 83.1 \\
    DynaMixer-L & MLP & $224^{2}$ & 97M & 27.4G &  \textbf{84.3}\\
    \bottomrule
    \end{tabular}}
    \caption{Top-1 accuracy comparisons with CNN-based models and  transformer based models on the ImageNet-1K image classification benchmark. Note that all models are trained without external data.  The ``Scale'' denotes the input size for training and testing stages. 
    The ``Param'' and ``FLOPs'' columns denote the number of parameters and the number of floating point operations, respectively.
    The values in the column Family, \textit{e.g.,} ``CNN'',  ``Trans'', ``Hybrid'', and ``FFT'', mean CNN-based, transformer-based, CNN-and-transformer-based, and fast Fourier transform based models. 
    }
    \label{tab:acc_all}
\end{table}

\paragraph{Model Latency Comparison.} We test the throughput of our model, ViP, and ResMLP with a batch size set of 32 on a single NVIDIA V100 GPU. Table~\ref{tab:acc_throughput} shows these results. We can see that among models achieving similar top-1 accuracies, our model achieves the best throughput. For example, DynaMixer-S and ViP-M/7 have similar top-1 performance. The throughput of our model is 551 images/s, while the throughput of ViP-M/7 is only 390 images/s. Moreover, DynaMixer-S is also faster than ResMLP B24 with a higher top-1 accuracy. This indicates that compared to SOTA MLP-like models, our model achieves similar or better performance with fewer parameters.

\begin{table}[h]
    \resizebox{.99\columnwidth}{!}{
    \centering
    \begin{tabular}{l|c|c|c|c}
    \toprule
    Model & \# Param & FLOPs & \begin{tabular}[c]{@{}c@{}}Throughput\\ image/s\end{tabular}  & Top-1 (\%)\\
    \midrule
    ResMLP S24 & 30M & 6.0G & 680 &79.4 \\
    ResMLP B24 & 116M & 23.0G & 215 &81.0 \\
    \midrule
    ViP-M/7 & 55M & 16.3G & 390 & 82.7\\
    \textbf{DynaMixer-S} &  \textbf{26M} &  \textbf{7.3G} & \textbf{551} & 82.7\\
    \midrule
    ViP-L/7 & 88M & 24.4G & 276 & 83.2\\
    \textbf{DynaMixer-M} & \textbf{57M} &  \textbf{17.0G} & \textbf{326} & 83.7\\
    \bottomrule
    \end{tabular}
    }
    \caption{The number of parameters, FLOPs and throughput of different models.} \label{tab:acc_throughput}
\end{table}

\paragraph{Limitations.} The numbers of parameters and FLOPs of DynaMixer are slightly larger than those of other models, as the mixing matrix is generated dynamically. In practice, such an increment over model size will affect the training time. 
Compared with ViPs of a similar number of parameters, the FLOPs of our models are slightly bigger and the throughputs are about $16\% \sim 18\%$ lower.
Moreover, the input image size for our model should be fixed, which restricts its applications on some downstream tasks, \textit{e.g.},  object detection, and segmentation. 
{
}

\subsection{Ablation Study}
In this subsection, we present ablation study results with DynaMixer-S to provide a deeper understanding of our methods.

 \paragraph{Ablation on the number of segments $S$.} To improve the robustness and generalization ability of our network, we introduce a multi-segment fusion mechanism in the DynaMixer operation. To show the effects of the number of segments, the performances with different values of segment dimensionalities, which is $\frac{D}{S}$, are reported in Table~\ref{tab:acc_s}. Considering the computational complexity, we only test 4 values of segment dimensionalities, which are $D$, 96, 48, and 24. 
 We can see that even without multiple segments, our model can still obtain satisfying accuracy ($82.2\%$). As $S$ increases ($\frac{D}{S}$ decreases), the performance becomes better. Moreover, our network is not sensitive to segment dimensionalities in a quite large range. 
\begin{table}[ht]
    \centering
    \begin{tabular}{l|c|c|c}
    \toprule 
    Model & $D/S$ & \# parameters & Top-1 $(\%)$ \\
    \midrule
    DynaMixer-S & $D$ & 26M & 82.2\\
    DynaMixer-S & 96 & 26M &82.4 \\
    DynaMixer-S & 48 & 26M &82.5 \\
    DynaMixer-S & 24 & 26M & 82.7 \\
    \bottomrule
    \end{tabular}
    \caption{The top-1 accuracies with different segment dimensionalities.}
    \label{tab:acc_s}
\end{table}

\paragraph{Ablation on weight generation methods.}
We study the effects of different methods to generate mixing matrices in this paragraph. The methods include ``Synthesizer (Random)'' and ``Synthesizer (Dense)'', which are described in \cite{synthesizer}. Once the mixing matrices are generated, they are applied to mixing tokens by following the procedure of DynaMixer, which means that tokens are mixed row by row and column by column with the multi-segment fusion mechanism. The mixing matrices generated by ``Synthesizer (Random)'' are actually trainable parameters. Thus it is similar to ViP~\cite{vip}, except that the information fusion among segments of channel is different. The mixing matrix generated by ``Synthesizer (Dense)'' does not consider the effects of other tokens. Thus, the mixing weight vector for a specific output token in ``Synthesizer (Dense)'' is generated merely with the information from the token at the same position. The performances of these methods are listed in Table~\ref{tab:as_diff_w}. 
We can see that ``Synthesize (Random)'' achieved an accuracy of 81.5\%, which is close to ViP's. It is natural since their mixing matrices are generated in the same way. The accuracy achieved by ``Synthesizer (Dense)'', which is 20\% larger than DynaMixer-S, is only 81.4\%, which demonstrates that leveraging the contents of all tokens is necessary, and the technique of reducing the computational complexity adopted in our model is effective.

\begin{table}[ht]
\resizebox{.99\columnwidth}{!}{
    \centering
    \begin{tabular}{l|c|c|c}
    \toprule
    Model & generating methods & \#  parameters & Top-1 $(\%)$ \\
    \midrule 
    DynaMixer-S & Synthesize(Random) & 25M & 81.5\\
    DynaMixer-S & Synthesizer(Dense) & 32M & 81.4\\
    DynaMixer-S & DynaMixer-S & 26M & 82.7\\
    \bottomrule
    \end{tabular}
    }
    \caption{The effects of different mixing matrix generation methods.}
    \label{tab:as_diff_w}
\end{table}

\paragraph{Ablation on the reduced dimensionality $d$.} In our model, dimensionality is reduced to $d$ by a liner projection to decrease the number of parameters. 
Ideally, $d$ should be large to provide enough information. However, we found that \textit{$d$ can be extremely low in practice}. The top-1 accuracies with different values of $d$ are shown in Table~\ref{tab:acc_diff_d}. We can see that our model with $d=1$ can still achieve an accuracy of 82.4\%, which is an absolute improvement of 0.9\% compared with ViP. Thus, even dimensionality is reduced to 1, the reduced token feature could provide information of all tokens to some extent.
The performances with values of $d$ set to 2, 4, and 8 are similar. Thus, $d$ is set to 
2 for DynaMixer-S in our experiments.
\begin{table}[ht]
    \centering
    \begin{tabular}{l|c|c|c}
    \toprule
    Model & $d$ & \# parameters & Top-1 $(\%)$ \\
    \midrule
    DynaMixer-S & 1 & 26M & 82.4 \\
    DynaMixer-S & 2 & 26M & 82.7 \\
    DynaMixer-S & 4 & 27M & 82.6 \\
    DynaMixer-S & 8 & 29M & 82.7 \\
    \bottomrule
    \end{tabular}
    \caption{The effects of reduced dimensionality.}
    \label{tab:acc_diff_d}
\end{table}

\paragraph{Ablation on the designs of DynaMixer Block.} 
In the DynaMixer block, the final results are from three components: row mixing,  column mixing, and channel mixing. The final results are a weighted sum of the above three. In this paragraph, we study the effects of row mixing, column mixing, channel mixing, and reweighting by removing one of them in turn from our model while keeping the other components unchanged. The results are shown in Table~\ref{tab:acc_r_c_c_r}. We can see that without row mixing or column mixing, the performance will decrease from 82.7\% to about 79\%. 
Moreover, if the channel mixing component is removed, the performance will drop 0.6\% absolutely (from 82.7\% to 82.1\%). Without the reweighting step, the performance will also become worse. Thus, all the components in our DynaMixer block are necessary for our model. Finally, we evaluate the performance of our model when the parameters of DynaMixer operations for row mixing and column mixing are shared, which is a straightforward way to reduce the number of parameters. We can see that the accuracy only drops to 82.2\%, which is still better than the other vision MLPs of similar size.

\begin{table}[ht]
    \centering
    \begin{tabular}{l|c|c}
    \toprule
    Model & \# parameters & Top-1 $(\%)$ \\
    \midrule
    DynaMixer-S & 26M & 82.7\\
    $\ $ - column mixing & 23M & 79.2 \\
    $\ $ - row mixing & 23M &79.4 \\
    $\ $ - channel fusion & 23M & 82.1\\ 
    $\ $ - reweighting &24M & 81.9 \\
    $\ $ + sharing DynaMixer op  &23M & 82.2 \\
    \bottomrule
    \end{tabular}
    \caption{The effects of different components in the DynaMixer block.}
    \label{tab:acc_r_c_c_r}
\end{table}
\subsection{Transfer Learning}
In this subsection, we show the advantages of our models on the transfer learning tasks. We transfer our pre-trained DynaMixer-S to
downstream datasets such as CIFAR10 and CIFAR100. We finetune our Dynamixer-S with 60 epochs by using the SGD optimizer and cosine learning rate decay. The results are given in Table~\ref{tab:cifar}. We find that our DynaMixer can achieve higher accuracies than ViP and the original ViT with similar model sizes on the downstream datasets.
\begin{table}[ht]
\resizebox{.95\columnwidth}{!}{
    \centering
    \begin{tabular}{l|c|c|c}
    \toprule
    Model & dataset & \# parameters & Top-1 $(\%)$ \\
    \midrule 
    ViT-S/16~\cite{vit} & CIFAR-10 & 49M & 97.1\\
    T2T-14~\cite{Yuan_2021_ICCV} &CIFAR-10 &22M & 97.5\\
    ViP-S/7~\cite{vip} &CIFAR-10 & 25M & 98.0\\
    DynaMixer-S & CIFAR-10 & 26M & \textbf{98.2}\\
    \midrule
    ViT-S/16~\cite{vit} & CIFAR-100 & 49M & 87.1\\
    T2T-14~\cite{Yuan_2021_ICCV} &CIFAR-100 &22M &88.4 \\
    ViP-S/7~\cite{vip} & CIFAR-100 & 25M & 88.4\\
    DynaMixer-S & CIFAR-100 & 26M & \textbf{88.6}\\
    \bottomrule
    \end{tabular}
    }
    \caption{The results of fine-tuning the pre-trained DynaMixer on ImageNet to downstream datasets: CIFAR10 and CIFAR100.}
    \label{tab:cifar}
\end{table}

\section{Conclusion}
MLP-based networks have recently been designed for vision recognition, targeting at introducing less inductive biases. The performances of MLP-based networks have been shown better than or comparable with transformer-based models. Existing MLP-based vision models utilize static trainable mixing matrices to perform token information fusion. In this paper, we proposed a novel MLP-based vision architecture, called DynaMixer, which further improves the visual recognition performance. In DynaMixer, the mixing matrices are generated dynamically by leveraging the contents of all the tokens to be mixed. To reduce the time complexity and meanwhile improve the robustness, we exerted a dimensionality reduction technique and a multi-segment fusion mechanism. Our DynaMixer model has been demonstrated to achieve state-of-the-art performance on the ImageNet recognition task. We also provided an insightful analysis for gaining a deeper understanding of our model.

\bibliography{dynamixer_bib}
\bibliographystyle{icml2022}



\end{document}